\documentclass[11pt]{article}

\usepackage[preprint]{acl}
\usepackage{times}
\usepackage{latexsym}
\usepackage[T1]{fontenc}
\usepackage[utf8]{inputenc}
\usepackage{microtype}
\usepackage{graphicx}
\usepackage{booktabs}
\usepackage{amsmath}
\usepackage{amssymb}
\usepackage{float}

\newcommand{\beps}{\boldsymbol{\epsilon}}

\title{Asynchrony Is Nearly Free for Evolution Strategies on Long-Horizon LLM Agentic Tasks} 

\author{
  William Hoy$^{1}$ \quad
  Jingxuan Fan$^{2}$ \quad
  Nurcin Celik$^{1}$ \quad
  Xu Pan$^{2}$ \\[0.5em]
  $^{1}$University of Miami \quad
  $^{2}$Harvard University \\[0.25em]
  \texttt{wjh58@miami.edu}, \texttt{jfan@g.harvard.edu} \\
  \texttt{celik@miami.edu}, \texttt{xupan@fas.harvard.edu}
}
\date{}

\begin{document}
\maketitle

\begin{abstract}

Synchronous policy updates can limit the efficiency of post-training for
long-horizon LLM agents. When trajectories span multiple interaction turns,
completion times can vary substantially, and synchronous update barriers
leave faster workers idling, wasting compute resources. Asynchronous
reinforcement learning addresses this inefficiency but introduces policy
lag and off-policy optimization. Evolution strategies (ES) are a
backpropagation-free alternative to gradient-based training, yet require
many perturbed-policy evaluations and have remained largely synchronous in
current LLM post-training.
We introduce bounded-staleness
asynchronous ES and investigate its training behavior under stale fitness measurements. On WebShop, we
systematically vary the lag and fraction of stale fitness measurements
and additionally introduce two stress tests to probe extreme
staleness cases. Natural asynchrony closely tracks synchronous ES, and
degradation in evaluation accuracy remains small even under extreme
staleness, including when all fitness measurements are stale. These findings suggest that ES can tolerate substantial policy staleness without explicit off-policy correction, supporting its use in asynchronous post-training for long-horizon LLM agents.


\end{abstract}

\section{Introduction}

Post-training language agents increasingly requires long-horizon interactions
with stateful environments. In such setups, trajectory lengths and completion times can vary widely across rollouts. Some rollouts finish quickly, while others hang for substantially longer. Synchronous training waits for the entire
update cohort to finish, leaving faster workers idle near the rollout barrier
and reducing hardware utilization
\citep{fu2025areal,lu2025rollflash,hou2026sao}.

Asynchronous reinforcement learning relaxes this barrier by consuming
completed trajectories while other rollouts remain in flight
\citep{nair2015massively,mnih2016a3c,espeholt2018impala}. Recent systems bring
this approach to language-model post-training
\citep{noukhovitch2024asyncrlhf,han2025asyncflow,fu2025areal,gao2025beyond}.
However, asynchronous training introduces policy lag: a trajectory can be generated by an
earlier policy and consumed after the learner has advanced. Increasing this
mismatch can compromise optimization stability
\citep{hou2026sao,zheng2025prosperity}, creating a tradeoff between worker
utilization and the usefulness of stale experience.

\begin{figure*}[t]
\centering
\includegraphics[width=\textwidth]
{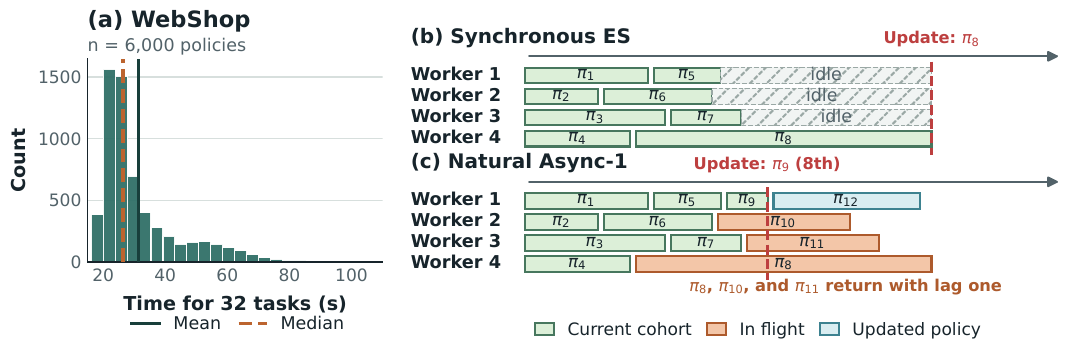}
\caption{WebShop rollout workload and asynchronous dispatch.
Left: rollout-batch durations for synchronous ES with 32 tasks
per perturbation, pooled across training.
Right: synchronous and asynchronous scheduling with four workers
and an eight-result update cohort. Asynchronous ES updates after
eight accepted evaluations complete while other evaluations
remain in flight.}
\label{fig:motivation-method}
\end{figure*}

Evolution strategies (ES) offer a backpropagation-free alternative to
policy-gradient post-training
\citep{qiu2026esatscale,hoy2026geometry}. ES evaluates parameter perturbations
using scalar trajectory rewards and avoids storing activations for
backpropagation. Recent work extends this approach to long-horizon language
agents \citep{zheng2026esopt}, but retains synchronous population updates.
Each perturbed policy is evaluated on a batch of tasks, so an ES cohort
combines many rollouts with potentially uneven completion times.
Figure~\ref{fig:motivation-method} illustrates this workload variation
and the barrier that asynchronous dispatch relaxes.

This raises a central question: \emph{how much policy staleness can ES
tolerate before delayed evaluation begins to degrade learning?} 
Unlike decoupled asynchronous policy-gradient methods, where the trajectories come from an older policy but the gradients are computed using its current model parameters, ES consumes fitness measurements of
perturbations around older central policies and directly applies fitness-weighted old perturbations to the current model parameters. A search direction that was useful at dispatch may therefore become less informative as the central policy evolves.
Our study therefore examines how the age of fitness measurements and the fraction of stale measurements within each cohort affect ES training effectiveness on LLM agentic tasks.

We introduce bounded-staleness asynchronous ES and use it on the
WebShop task \citep{yao2022webshop}, where agents search for products, inspect
options, and complete purchases through multi-turn interactions. 
We compare synchronous ES with naturally asynchronous ES and controlled asynchronous variants with prescribed fractions and ages of stale fitness measurements. Across these settings, we observe only minor differences in evaluation accuracy. To probe the effects of more extreme staleness, we introduce two stress test setups with extreme staleness, i.e., ``long-delay test'' and ``staircase test''. Even under these extreme conditions, we observe only small reductions in evaluation accuracy relative to synchronous ES.

Our contributions are:
\begin{enumerate}
    \item A bounded-staleness asynchronous ES algorithm that continuously
    dispatches perturbation evaluations, tracks policy versions, and updates
    the central policy using accepted fitness measurements without explicit
    off-policy correction.

    \item A systematic study of policy lag and the fraction of stale fitness
    measurements on WebShop, showing only minor performance degradation
    relative to synchronous ES across the tested settings.

\end{enumerate}

\section{Asynchronous Evolution Strategies}
\label{sec:method}

Synchronous ES evaluates $N$ perturbations around the same central
policy $\theta_t$ and waits for the complete cohort before updating.
Uneven rollout-batch durations can therefore leave faster workers
idle near the update barrier
(Figure~\ref{fig:motivation-method}).

Our bounded-staleness asynchronous ES algorithm dispatches evaluations
across $G$ GPU workers, each evaluating one perturbed policy at a
time. Each evaluation returns its perturbation seed, mean task-batch
fitness $R_i$, and dispatch policy version $v_i$:
\begin{equation}
 R_i = R(\theta_{v_i}+\sigma\beps_i),
 \label{eq:async_reward}
\end{equation}
where $\sigma$ is the noise scale and $\beps_i$ is reconstructed
from the seed. At central-policy version $t$, the result has policy
lag $\ell_i=t-v_i$. Results exceeding the configured bound
$\ell_{\max}$ are discarded.

Once $N$ admissible results are available, the coordinator
standardizes their fitness measurements,
$Z_i=(R_i-\mu_R)/(\sigma_R+10^{-8})$, using the accepted cohort's
mean and standard deviation, and applies
\begin{equation}
 \theta_{t+1}
 = \theta_t+\frac{\alpha}{N}\sum_{i=1}^{N}Z_i\beps_i,
 \label{eq:es-update}
\end{equation}
where $\alpha$ is the ES step size. Evaluations from different
policy versions contribute their original perturbation directions
to the current policy without explicit off-policy correction.
Workers incorporate committed updates between evaluations and
immediately begin new perturbations. Async-1 uses
$\ell_{\max}=1$.


\section{Experimental Setup}
\label{sec:setup}

We use Qwen3-4B-Instruct-2507 and present WebShop
experiments with $N=16$ perturbations and $D=16$ tasks per
perturbation for 200 ES updates in the main text and provide results for additional configurations
of $30\times32$ for 200 updates and $30\times100$ for
120 updates in the Appendix~\ref{app:webshop}, where $N\times D$ denotes perturbations per
update and tasks per perturbation. 

ES uses full-parameter Gaussian perturbations with noise
scale $\sigma=0.0015$, step size $\alpha=0.00075$, and 4
GPU workers. Fitness combines partial task score with an
exact-success bonus. We report exact task success as the
primary metric.

Training excludes a fixed 128-task eval set, evaluated
every ten updates at temperature 0.4. We compare synchronous ES,
natural Async-1, controlled lags,
long-delay test, and the staircase test.

GRPO provides a gradient-based reference, using 16 tasks
and 16 responses per task for 100 updates. Asynchronous
GRPO uses token-level truncated importance sampling.
Complete setup appears in
Appendix~\ref{app:config}.

\section{Results}
\label{sec:results}

\subsection{Policy lag and stale fraction}

\begin{figure}[t]
 \centering
 \includegraphics[width=\columnwidth]
 {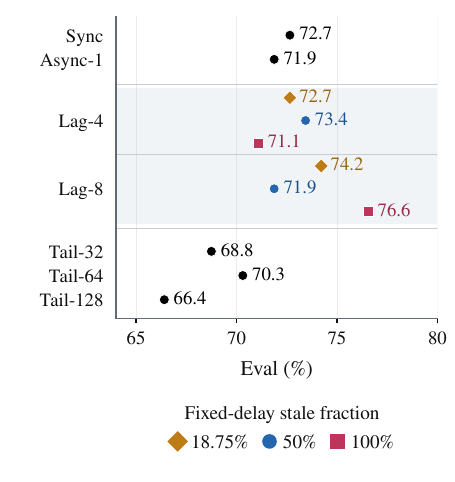}
 \caption{WebShop $16\times16$ exact success on the fixed
 128-task eval set at update 200. Controlled Lag-4/8
 sub-rows are ordered by stale fraction: 18.75\%, 50\%,
 and 100\%. Tail conditions mix fresh and delayed
 evaluations. Each point represents one training run.}
 \label{fig:webshop-final}
\end{figure}
We separately control the policy lag (i.e. age) and fraction of stale
fitness measurements within each update cohort to isolate their effect.
Fixed-delay conditions combine fresh policies with a prescribed
fraction of stale policies delayed by $L$ updates. 
Fresh and stale fitness measurements share the same cohort-wide
normalization, rather than being normalized separately by policy
version.

Figure~\ref{fig:webshop-final} summarizes final exact
success for $16\times16$. Async-1 finishes at
71.9\%, 0.8 percentage points below synchronous ES at 72.7\%.

Increasing either policy lag or the fraction of stale fitness measurements does not consistently degrade performance. In the $16\times16$ Lag-8 configuration, entirely stale cohorts outperform partially stale ones, although these differences should be interpreted cautiously given one run per condition. One possible explanation concerns cohort-wide fitness normalization. As training progresses, newer source policies may have higher mean fitness than older ones. Mixing their evaluations therefore introduces between-policy fitness differences into the perturbation weights, potentially increasing finite-population update noise and changing the effective update scale. Cohorts evaluated around a single stale policy avoid this source of heterogeneity, which may help explain their stronger performance relative to mixed cohorts. Future studies are needed to test this hypothesis.


Natural Async-1 GRPO finishes at 82.8\% and synchronous
GRPO at 77.3\% on the same monitoring tasks, exceeding
the ES endpoints.
Complete GRPO curves and controlled-delay results appear
in Appendix~\ref{app:webshop-grpo}. 

\subsection{Long-delay test}

The long-delay tests combine
several policy ages in one cohort. Half of the policies are fresh; half of the policies have a mixture of ages up to
32, 64, or 128 updates which we call the ``tail number'' (see Appendix Table~\ref{tab:lag-mixtures} for the exact distribution). Tail-128 finishes at 66.4\%,
compared with 72.7\% for synchronous ES. Full curves, peak and final scores, short-delay controls,
and the supporting $30\times32$ and $30\times100$
results appear in Appendix~\ref{app:webshop}.

\subsection{Staircase test}

\begin{figure}[t]
 \centering
 \includegraphics[width=\columnwidth]
 {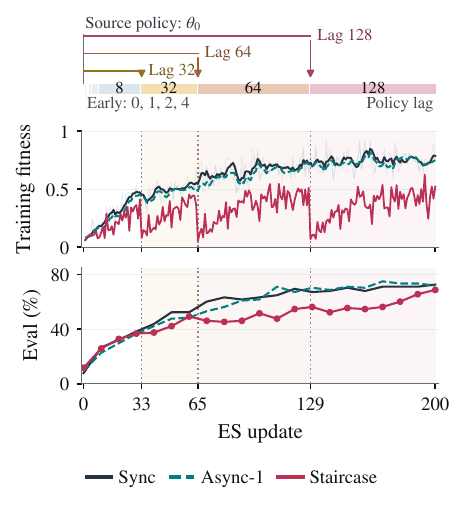}
 \caption{WebShop $16\times16$ staircase test.
 Brackets link evaluations around $\theta_0$ to consumption
 at updates 33, 65, and 129, with lags 32, 64, and 128.
 Staircase fitness is raw; reference fitness shows
 five-update trailing means with faint raw values.}
 \label{fig:webshop-staircase}
\end{figure}
The staircase test selects the largest
available lag from $\{1,2,4,8,32,64,128\}$. Its first cohort is fresh,
and every subsequent cohort is entirely stale with the largest
available lag. 
Figure~\ref{fig:webshop-staircase} illustrates this process. At each lag transition, the source policy resets to the initial checkpoint, causing a sharp drop in the reported fitness. Between transitions, the source-policy version advances, so the subsequent recovery reflects improvements in earlier source policies rather than the current policy.

Even in this extreme case of staleness, the evaluation accuracy still shows a steady increase, finishing only 3.9 percentage points below synchronous ES and 3.1 points below Async-1 at update 200. The $30\times32$ staircase
and its matched references appear in
Appendix~\ref{app:webshop}.  Together with the results in the early sections, these findings suggest that ES can tolerate substantial policy staleness without explicit off-policy correction.




\subsection{Capability retention}

\begin{table}[h]
\centering
\small
\begin{tabular}{lrr}
\toprule
Method & MMLU-Pro & HellaSwag \\
\midrule
Base
  & 0.5894 & 0.8018 \\
Synchronous ES
  & 0.5586 & 0.7603 \\
Natural Async-1 ES
  & 0.5410 & 0.7559 \\
Lag-4 (18.75\% stale)
  & 0.5293 & 0.7349 \\
Lag-8 (18.75\% stale)
  & 0.5181 & 0.7651 \\
Lag-4 (50\% stale)
  & 0.5371 & 0.7695 \\
Lag-4 (100\% stale)
  & 0.5542 & 0.7539 \\
Lag-8 (50\% stale)
  & 0.4927 & 0.7280 \\
Lag-8 (100\% stale)
  & 0.5381 & 0.7476 \\
Tail-32
  & 0.5259 & 0.7178 \\
Tail-64
  & 0.5620 & 0.6816 \\
Tail-128
  & 0.5400 & 0.6665 \\
Staircase
  & 0.5557 & 0.7256 \\
\bottomrule
\end{tabular}
\caption{General-capability retention after WebShop post-training
with ES ($16\times16$), initialized from Qwen3-4B-Instruct-2507.
All methods are evaluated after 200 updates.
Base-model accuracies are 0.5894 on MMLU-Pro and 0.8018 on HellaSwag.
Each final checkpoint uses the same 2,048 examples per benchmark
and greedy decoding.}
\label{tab:forgetting-webshop-16x16}
\end{table}

We evaluate final WebShop checkpoints on fixed sets of 2,048 examples
from MMLU-Pro and HellaSwag using greedy decoding.
Table~\ref{tab:forgetting-webshop-16x16} reports the $16\times16$
results after 200 updates. Synchronous ES reduces accuracy relative
to the base model by 3.1 percentage points on MMLU-Pro and 4.2 on
HellaSwag. Natural Async-1 shows additional reductions of 1.8 and
0.4 points, respectively.

The additional $30\times32$ and $30\times100$ configurations show
smaller median declines on both benchmarks, while GRPO remains
within 0.9 points of base-model accuracy on MMLU-Pro and 0.5 on
HellaSwag (Appendix~\ref{app:retention}, Table~\ref{tab:forgetting-webshop}).

\subsection{Compute comparison}
ES evaluates perturbed policies using inference only,
avoiding backpropagation and reference-model inference.
However, its total compute depends on the number of
perturbations and tasks evaluated. Under our approximation,
the $16\times16$, $30\times32$, and $30\times100$ ES
runs use $0.40\times$, $1.50\times$, and $2.81\times$
the model-compute budget of the 100-update GRPO run,
respectively. Accounting details appear in
Appendix~\ref{app:compute_matching}.

\section{Discussion and Conclusion}

ES is emerging as a backpropagation-free alternative for LLM
post-training \citep{qiu2026esatscale,hoy2026geometry,zheng2026esopt}.
Its reliance on scalar rewards rather than gradients makes it
particularly attractive for long-horizon agentic tasks, where
performance is evaluated through interactions with external
environments. However, synchronous population updates can limit
its efficiency when rollout completion times vary substantially.

Our results suggest that strict synchronization is not necessary
for effective ES training. Natural asynchronous ES closely tracks
synchronous ES, and even entirely stale cohorts can sustain learning
without explicit off-policy correction. 

These findings support asynchronous ES as a promising direction
for scalable LLM-agent post-training. Nevertheless, extreme delays
can reduce task performance, and our capability-retention experiments
reveal broader performance degradation even when WebShop learning
remains effective. Future work should investigate the mechanisms
underlying staleness tolerance, improve capability preservation,
and evaluate practical efficiency across models
and agentic environments.



\clearpage
\section*{Limitations}
\label{sec:limitations}

Our experiments cover one model and one environment,
with one training run per condition. Replicated runs
are needed to assess variation across training seeds. Controlled delays test optimization under staleness,
not real asynchronous resource contention. 

\bibliography{reference}

\clearpage
\appendix

\section{Potential Risks}

Our study focuses on the general applicability of the asynchronous ES algorithm. The algorithm does not raise specific risks other than the common risks of LLM post-training, such as misalignment. 

\section{LLM usage}

LLM tools assisted with experiment code, manuscript wording, and LaTeX formatting.

\section{Code availability}

The asynchronous ES trainer is publicly available at
\url{https://github.com/Bhoy1/async-es}.
We plan to add further experiment-specific scripts, configurations,
and result artifacts for the experiments reported in this work,
including the WebShop and controlled-staleness studies, in upcoming
updates. The repository will be maintained as an active research
toolbox, with additional parameterizations, benchmarks, and model
configurations planned.

Parts of the implementation build on ES at Scale
(\url{https://github.com/VsonicV/es-at-scale}), which is distributed
under the Academic Public License. This license permits free
noncommercial use, including academic research; commercial use
requires a separate commercial license.

\section{Complete Experimental Configurations}
\label{app:config}

\subsection{WebShop}

\paragraph{Evaluation.}
We reserve task IDs 0--127 for monitoring and exclude them from training. We evaluate the current central ES policy or GRPO policy before training and every ten updates on the same 128 monitoring tasks. Each evaluation generates one trajectory per task at temperature 0.4.

\paragraph{Rewards and rollout settings.}
Let $s\in[0,1]$ denote the environment's partial task score and
$c\in\{0,1\}$ denote exact task success. Both methods use the
trajectory reward
\begin{equation}
 r = 2s + 8c.
\end{equation}
ES fitness is the mean reward across a perturbation's task batch.
The primary reported task metric is the mean of $c$, rather
than the fraction of trajectories receiving positive partial
reward. For visualization, training fitness and reward are normalized
to $[0,1]$ by dividing by 10.

Rollouts permit at most 15 interaction turns and 1,024 generated
tokens per turn. Training uses temperature 1.0 and top-$p=1.0$ without top-$k$ truncation.
The model context limit is 32,768 tokens. 

\paragraph{GPU configuration.}
Each training run uses four GPUs. ES assigns one
perturbation-evaluation worker to each GPU. GRPO uses
two GPUs for policy optimization and reference-model
inference, and two for rollout generation.

\paragraph{GRPO hyperparameters.}
GRPO uses 16 prompts and 16 responses per prompt, producing
256 trajectories per update, with one optimization epoch per
batch. The optimizer uses learning rate $10^{-6}$, Adam
coefficients $(0.9,0.999)$, weight decay 0.01, and gradient
clipping at 1.0. PPO ratio clipping is 0.2 in both directions.
A loss-side $k_3$ KL penalty has coefficient 0.01; no entropy
loss is applied. Token-level truncated importance sampling clips the correction ratio at 2.0. Async-1 permits at most one update of policy lag.

\paragraph{ES hyperparameters.}
All reported ES configurations perturb the full model with
Gaussian noise scale $\sigma=0.0015$ and use ES step size
$\alpha=0.00075$. The training seed is 42 and the WebShop
environment seed is 0. Four GPU workers independently evaluate
perturbations. The $16\times16$ and $30\times32$ configurations run for 200 updates; $30\times100$ runs for 120 updates. Natural Async-1 consumes one-update-old evaluations
for 18.56\%, 9.85\%, and 9.83\% of accepted results
at $16\times16$, $30\times32$, and $30\times100$, measured across each full run.

\paragraph{Lag schedules.}
For $30\times32$, controlled Lag-4 and Lag-8 use stale counts
of 3, 15, or 30 per cohort, corresponding to 10\%, 50\%, and
100\%. For $16\times16$, controlled Lag-4 and Lag-8 use
3, 8, or 16 stale evaluations, corresponding to 18.75\%,
50\%, and 100\%. The $30\times100$ controlled conditions
use three stale evaluations per cohort at lag 4 or 8, corresponding to 10\%. The lag begins contributing once its requested policy age is available. Before that point, fresh evaluations fill the cohort.

\paragraph{Long-delay schedules.}
Table~\ref{tab:lag-mixtures} gives the exact composition of each
cohort after all requested lags become available. Each mixture
keeps half of the cohort fresh. Longer-tail conditions replace
some lag-8 evaluations with evaluations at lags 32, 64, or 128.
During startup, unavailable lag buckets are replaced by fresh
evaluations.

\begin{table*}[t]
\centering
\small
\begin{tabular}{llrrrrrrrr}
\toprule
Configuration & Schedule
& Fresh & Lag 1 & Lag 2 & Lag 4 & Lag 8
& Lag 32 & Lag 64 & Lag 128 \\
\midrule
$30\times32$ & Tail-32
  & 15 & 3 & 2 & 2 & 6 & 2 & 0 & 0 \\
             & Tail-64
  & 15 & 3 & 2 & 2 & 4 & 2 & 2 & 0 \\
             & Tail-128
  & 15 & 3 & 2 & 2 & 2 & 2 & 2 & 2 \\
\midrule
$16\times16$ & Tail-32
  & 8 & 2 & 1 & 1 & 3 & 1 & 0 & 0 \\
             & Tail-64
  & 8 & 2 & 1 & 1 & 2 & 1 & 1 & 0 \\
             & Tail-128
  & 8 & 2 & 1 & 1 & 1 & 1 & 1 & 1 \\
\bottomrule
\end{tabular}
\caption{Number of accepted perturbation evaluations at each
policy lag in the long-tail cohorts.}
\label{tab:lag-mixtures}
\end{table*}

\paragraph{Staircase test.}
At central-policy version $t$, the staircase selects
\begin{equation}
 \begin{aligned}
 \mathcal{L}&=\{1,2,4,8,32,64,128\},\\
 L(t)&=\max\{L\in\{0\}\cup\mathcal{L}:L\leq t\}.
 \end{aligned}
\end{equation}
All $N$ evaluations in the cohort have lag $L(t)$. Version
$t=0$ therefore uses a fresh cohort; every later cohort is
entirely stale. With updates numbered from one, lag transitions
occur at updates 2, 3, 5, 9, 33, 65, and 129. This schedule
changes lag alongside training stage and is interpreted as
a test rather than an isolated fixed-lag comparison.

\section{Additional WebShop Results}
\label{app:webshop}

We report 30 ES conditions across three population--task-batch
configurations and four GRPO conditions. Each condition represents
one training run. Figures show training fitness alongside current-policy exact success on
the fixed 128-task monitoring set. Non-staircase training curves use five-update trailing means;
staircase fitness is unsmoothed. Tables report the highest
observed scheduled success and final success, assigning tied
peaks to the earliest update. Long-tail mixture compositions appear in
Table~\ref{tab:lag-mixtures}; the annotated staircase analysis
appears in Figure~\ref{fig:webshop-staircase}.

\subsection{WebShop ES $16\times16$}

\begin{figure*}[t]
  \centering
  \includegraphics[width=\textwidth]
    {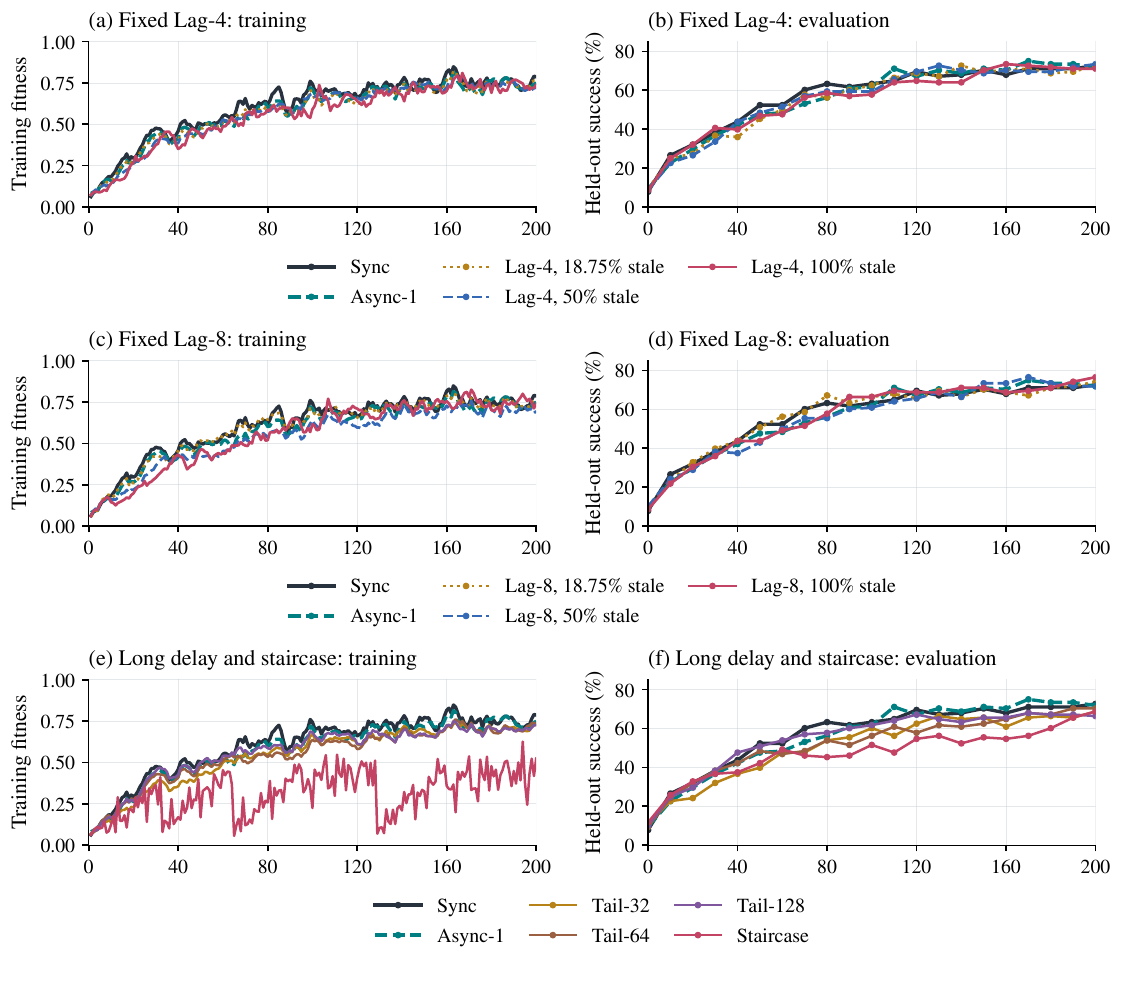}
  \caption{WebShop $16\times16$ ES learning curves
  through update 200. Left: training
  fitness. Right: eval success.
  Sync and Async-1 are repeated as references
  across schedule groups. Non-staircase fitness uses
  five-update trailing means; staircase fitness is raw.
  Each condition represents one training run.}
  \label{fig:webshop-p16-t16}
\end{figure*}

Figure~\ref{fig:webshop-p16-t16} shows the training and
monitoring curves. Table~\ref{tab:webshop-p16-t16}
summarizes peak and final success.

\begin{table*}[t]
\centering
\small
\begin{tabular}{lrrr}
\toprule
Condition & Peak success (\%) & Peak update & Final success (\%) \\
\midrule
Sync & 72.66 & 200 & 72.66 \\
Natural Async-1 & 75.00 & 170 & 71.88 \\
Lag-4, 18.75\% stale & 72.66 & 140 & 72.66 \\
Lag-4, 50\% stale & 73.44 & 200 & 73.44 \\
Lag-4, 100\% stale & 73.44 & 160 & 71.09 \\
Lag-8, 18.75\% stale & 74.22 & 200 & 74.22 \\
Lag-8, 50\% stale & 76.56 & 170 & 71.88 \\
Lag-8, 100\% stale & 76.56 & 200 & 76.56 \\
Tail-32 & 68.75 & 200 & 68.75 \\
Tail-64 & 70.31 & 190 & 70.31 \\
Tail-128 & 67.97 & 170 & 66.41 \\
Staircase & 68.75 & 200 & 68.75 \\
\bottomrule
\end{tabular}
\caption{WebShop $16\times16$ exact success on the fixed 128-task set. Peak is the highest logged scheduled evaluation over training, with ties assigned to the earliest update. Final is update 200. Each condition represents one training run.}
\label{tab:webshop-p16-t16}
\end{table*}

\subsection{WebShop ES $30\times32$}

\begin{figure*}[t]
  \centering
  \includegraphics[width=\textwidth]
    {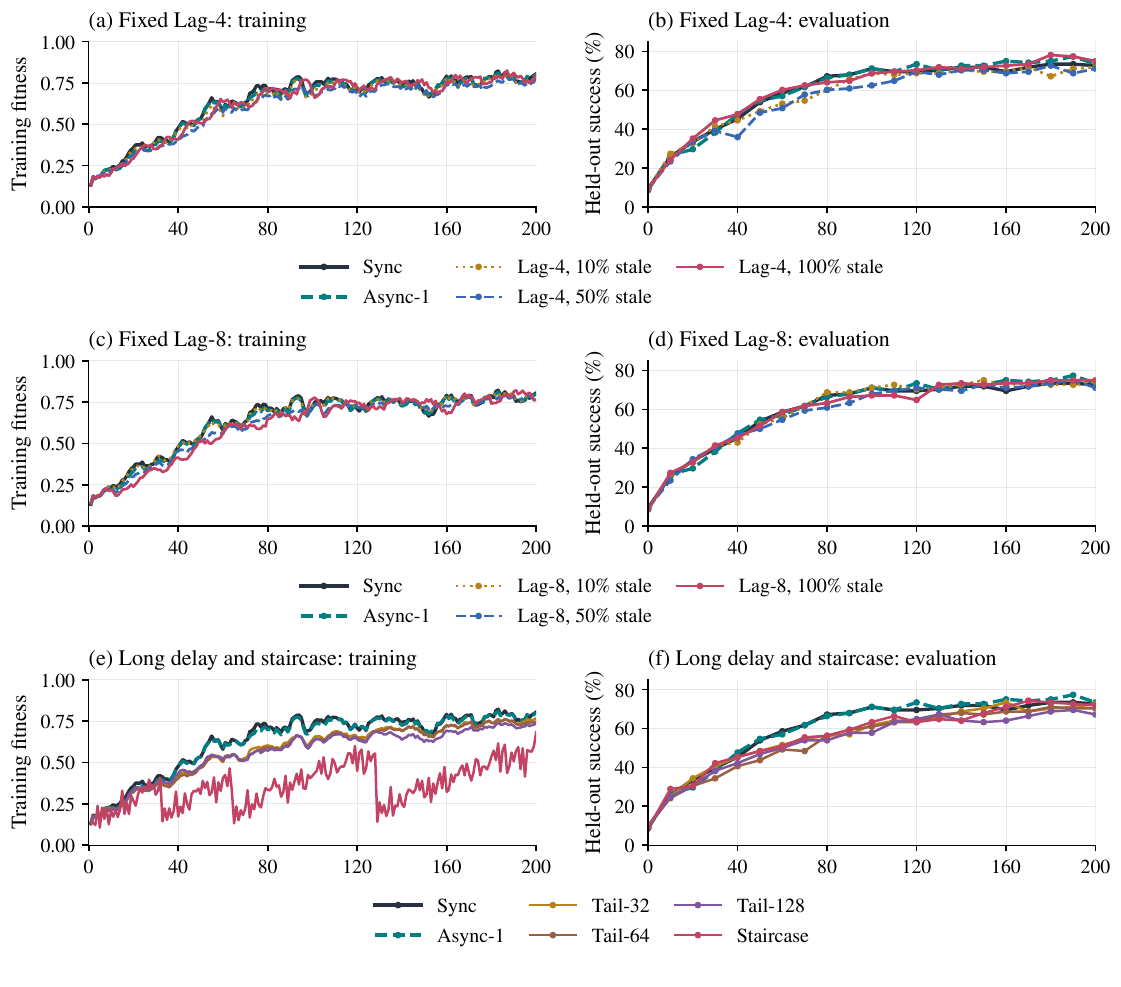}
  \caption{WebShop $30\times32$ ES learning curves
  through update 200. Left: training
  fitness. Right: eval success.
  Sync and Async-1 are repeated as references
  across schedule groups. Non-staircase fitness uses
  five-update trailing means; staircase fitness is raw.
  Each condition represents one training run.}
  \label{fig:webshop-p30-t32}
\end{figure*}

Figure~\ref{fig:webshop-p30-t32} shows the training and
monitoring curves. Table~\ref{tab:webshop-p30-t32}
summarizes peak and final success.

\begin{figure*}[t]
 \centering
 \includegraphics[width=\columnwidth]
 {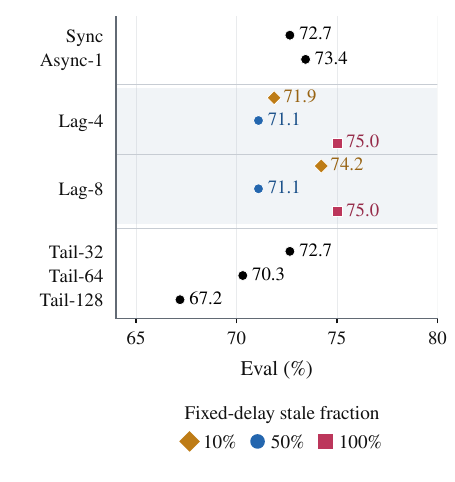}
 \caption{WebShop $30\times32$ exact success on the
 fixed 128-task eval set at update 200.
 Controlled Lag-4/8 sub-rows use 
 fractions of 10\%, 50\%, and 100\%.
 Tail conditions mix fresh and delayed
 evaluations. Each point represents one training run.}
 \label{fig:webshop-final-p30-t32}
\end{figure*}

\begin{figure*}[t]
 \centering
 \includegraphics[width=\columnwidth]
 {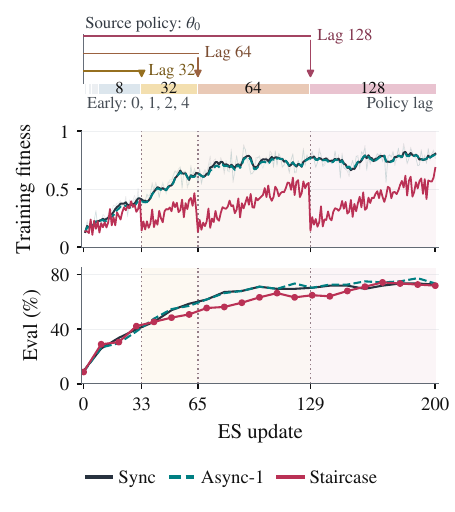}
 \caption{WebShop $30\times32$ staircase stress test.
 Brackets link evaluations around $\theta_0$ to consumption
 at updates 33, 65, and 129, with lags 32, 64, and 128.
 Staircase fitness is raw; reference fitness shows
 five-update trailing means with faint raw values.}
 \label{fig:webshop-staircase-p30-t32}
\end{figure*}

Figures~\ref{fig:webshop-final-p30-t32} and
\ref{fig:webshop-staircase-p30-t32} provide the
$30\times32$ counterparts to the main-text
$16\times16$ comparisons.

\begin{table*}[t]
\centering
\small
\begin{tabular}{lrrr}
\toprule
Condition & Peak success (\%) & Peak update & Final success (\%) \\
\midrule
Sync & 73.44 & 180 & 72.66 \\
Natural Async-1 & 77.34 & 190 & 73.44 \\
Lag-1, 50\% stale & 73.44 & 150 & 70.31 \\
Lag-1, 100\% stale & 75.00 & 190 & 73.44 \\
Lag-4, 10\% stale & 71.88 & 170 & 71.88 \\
Lag-4, 50\% stale & 72.66 & 180 & 71.09 \\
Lag-4, 100\% stale & 78.12 & 180 & 75.00 \\
Lag-8, 10\% stale & 75.00 & 150 & 74.22 \\
Lag-8, 50\% stale & 75.00 & 190 & 71.09 \\
Lag-8, 100\% stale & 75.00 & 180 & 75.00 \\
Tail-32 & 73.44 & 160 & 72.66 \\
Tail-64 & 71.09 & 180 & 70.31 \\
Tail-128 & 69.53 & 190 & 67.19 \\
Staircase & 74.22 & 170 & 71.88 \\
\bottomrule
\end{tabular}
\caption{WebShop $30\times32$ exact success on the fixed 128-task set. Peak is the highest logged scheduled evaluation over training, with ties assigned to the earliest update. Final is update 200. Each condition represents one training run.}
\label{tab:webshop-p30-t32}
\end{table*}

\subsection{WebShop ES $30\times100$}

\begin{figure*}[t]
  \centering
  \includegraphics[width=\textwidth]
    {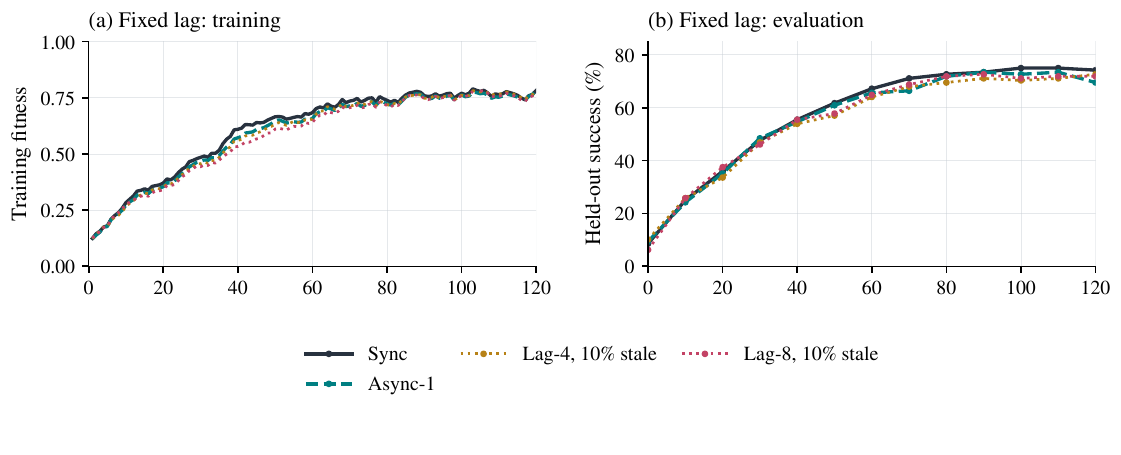}
\caption{WebShop $30\times100$ ES learning curves
through update 120. Left: training fitness Right:
eval success
Async-1 denotes natural asynchrony; controlled
Lag-4 and Lag-8 use 10\% stale evaluations. Each condition represents one training run.}
  \label{fig:webshop-p30-t100}
\end{figure*}

Figure~\ref{fig:webshop-p30-t100} shows the training
and monitoring curves.
Table~\ref{tab:webshop-p30-t100} summarizes peak
and final success. 

\begin{table*}[t]
\centering
\small
\begin{tabular}{lrrr}
\toprule
Condition & Peak success (\%) & Peak update & Final success (\%) \\
\midrule
Sync & 75.00 & 100 & 74.22 \\
Natural Async-1 & 73.44 & 90 & 69.53 \\
Lag-4, 10\% stale & 72.66 & 120 & 72.66 \\
Lag-8, 10\% stale & 72.66 & 90 & 71.88 \\
\bottomrule
\end{tabular}
\caption{WebShop $30\times100$ exact success on the fixed 128-task set. Peak is the highest logged scheduled evaluation over training, with ties assigned to the earliest update. Final is update 120. Each condition represents one training run.}
\label{tab:webshop-p30-t100}
\end{table*}

\subsection{WebShop GRPO}
\label{app:webshop-grpo}

\begin{figure*}[t]
  \centering
  \includegraphics[width=\textwidth]
    {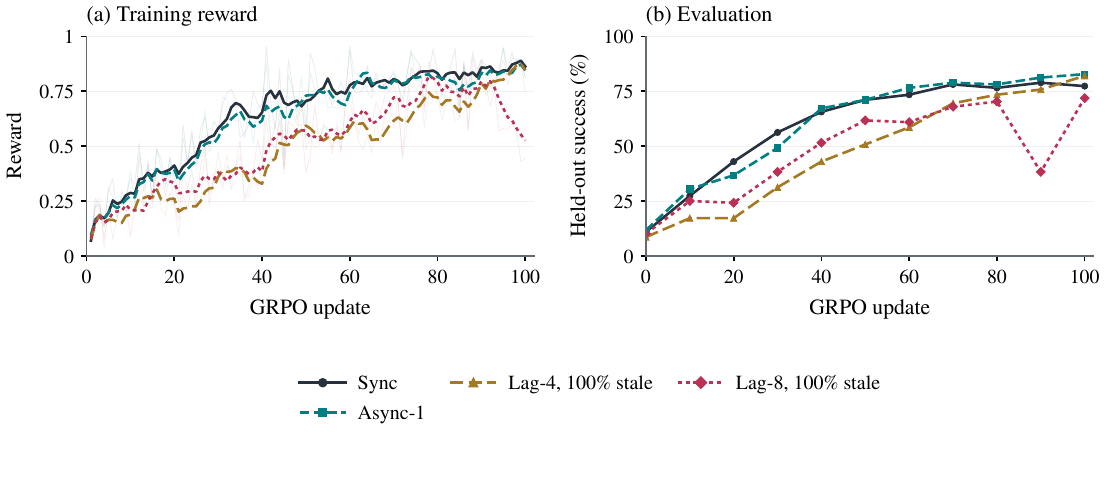}
  \caption{WebShop GRPO learning curves through
  update 100. Left: training reward Right: eval success
  Each condition represents one training run.}
  \label{fig:webshop-grpo}
\end{figure*}

Figure~\ref{fig:webshop-grpo} shows the complete GRPO
learning curves, and
Table~\ref{tab:webshop-grpo-peak-final} summarizes
peak and final exact success. All conditions use
16 tasks with 16 responses per task for 100 updates.
We compare synchronous GRPO, natural Async-1, and
controlled Lag-4 and Lag-8.

Natural Async-1 has an observed stale-group fraction
of 99\% across the full run. Controlled Lag-4 and
Lag-8 use entirely stale groups after warm-up,
giving whole-run stale fractions of 96\% and 92\%. 

\begin{table*}[t]
  \centering
  \small
  \begin{tabular}{lrrr}
    \toprule
    Condition & Peak success (\%) & Peak update
              & Final success (\%) \\
    \midrule
    Synchronous GRPO
      & 78.91 & 90 & 77.34 \\
    Natural Async-1 GRPO
      & 82.81 & 100 & 82.81 \\
    Controlled Lag-4 GRPO (100\% stale)
      & 82.03 & 100 & 82.03 \\
    Controlled Lag-8 GRPO (100\% stale)
      & 71.88 & 100 & 71.88 \\
    \bottomrule
  \end{tabular}
  \caption{WebShop GRPO exact success on the fixed
  128-task eval set. Peak is the highest
  success, with ties assigned
  to the earliest update; final is update 100.
  Each condition represents one training run.}
  \label{tab:webshop-grpo-peak-final}
\end{table*}

\section{General-Capability Retention}
\label{app:retention}

\subsection{Evaluation Configuration}
\label{app:forgetting_configuration}

We evaluate final WebShop checkpoints on the same fixed
2,048 examples from each of MMLU-Pro and HellaSwag,
selected using subset seed 2357. All checkpoints are
initialized from Qwen3-4B-Instruct-2507 \citep{yang2025qwen3}, which serves
as the base-model reference. Evaluation uses greedy
decoding at temperature 0 with a maximum generation
length of 1,024 tokens. Prompt formatting, demonstrations,
answer extraction, and scoring are held constant across
methods.

MMLU-Pro uses examples from its test split with five
category-matched demonstrations from the validation
split. HellaSwag uses validation examples without
demonstrations and asks the model to generate an option
label from 0 to 3. Accuracy is computed by extracting
the generated answer and comparing it with the reference
label. These are generative multiple-choice evaluations,
rather than log-likelihood rankings of candidate answers.

\subsection{Results}

Table~\ref{tab:forgetting-webshop} reports the base model,
four GRPO conditions, and 18 ES conditions. GRPO remains
near base-model accuracy, whereas all ES conditions
score below base on HellaSwag. MMLU-Pro changes vary
across configurations; Async-1 at $30\times100$
essentially matches base.

The $16\times16$ conditions generally show larger
declines, while $30\times100$ shows milder changes.
However, retention is not monotonically ordered by
policy lag or stale fraction.

\begin{table*}[t]
\centering
\footnotesize
\begin{tabular}{lrrrrr}
\toprule
Method & Updates & MMLU-Pro & $\Delta$
       & HellaSwag & $\Delta$ \\
\midrule
\multicolumn{6}{l}{\textit{Base and GRPO}} \\
Base
  & -- & 0.5894 & -- & 0.8018 & -- \\
Synchronous GRPO
  & 100 & 0.5815 & $-0.0078$ & 0.7988 & $-0.0029$ \\
Natural Async-1 GRPO
  & 100 & 0.5811 & $-0.0083$ & 0.7998 & $-0.0020$ \\
Controlled Lag-4 GRPO (100\% stale)
  & 100 & 0.5894 & $+0.0000$ & 0.7974 & $-0.0044$ \\
Controlled Lag-8 GRPO (100\% stale)
  & 100 & 0.5869 & $-0.0024$ & 0.8008 & $-0.0010$ \\

\midrule
\multicolumn{6}{l}{\textit{ES: $30\times32$}} \\
Synchronous ES
  & 200 & 0.5820 & $-0.0073$ & 0.7148 & $-0.0869$ \\
Natural Async-1 ES
  & 200 & 0.5703 & $-0.0190$ & 0.7466 & $-0.0552$ \\
Controlled Lag-4 (10\% stale)
  & 200 & 0.5679 & $-0.0215$ & 0.7524 & $-0.0493$ \\
Controlled Lag-8 (10\% stale)
  & 200 & 0.5840 & $-0.0054$ & 0.7598 & $-0.0420$ \\
Controlled Lag-1 (50\% stale)
  & 200 & 0.5723 & $-0.0171$ & 0.7529 & $-0.0488$ \\
Controlled Lag-1 (100\% stale)
  & 200 & 0.5703 & $-0.0190$ & 0.7412 & $-0.0605$ \\
Controlled Lag-4 (50\% stale)
  & 200 & 0.5527 & $-0.0366$ & 0.7505 & $-0.0513$ \\
Controlled Lag-4 (100\% stale)
  & 200 & 0.5845 & $-0.0049$ & 0.7085 & $-0.0933$ \\
Controlled Lag-8 (50\% stale)
  & 200 & 0.5771 & $-0.0122$ & 0.7354 & $-0.0664$ \\
Controlled Lag-8 (100\% stale)
  & 200 & 0.5605 & $-0.0288$ & 0.7622 & $-0.0396$ \\
Tail-32
  & 200 & 0.5659 & $-0.0234$ & 0.7407 & $-0.0610$ \\
Tail-64
  & 200 & 0.5640 & $-0.0254$ & 0.7334 & $-0.0684$ \\
Tail-128
  & 200 & 0.5781 & $-0.0112$ & 0.7598 & $-0.0420$ \\
Staircase
  & 200 & 0.5708 & $-0.0186$ & 0.7568 & $-0.0449$ \\

\midrule
\multicolumn{6}{l}{\textit{ES: $30\times100$}} \\
Synchronous ES
  & 120 & 0.5825 & $-0.0068$ & 0.7739 & $-0.0278$ \\
Natural Async-1 ES
  & 120 & 0.5898 & $+0.0005$ & 0.7891 & $-0.0127$ \\
Controlled Lag-4 (10\% stale)
  & 120 & 0.5864 & $-0.0029$ & 0.7827 & $-0.0190$ \\
Controlled Lag-8 (10\% stale)
  & 120 & 0.5786 & $-0.0107$ & 0.7822 & $-0.0195$ \\
\bottomrule
\end{tabular}
\caption{General-capability retention after WebShop post-training
with GRPO and ES ($30\times32$ and $30\times100$),
initialized from Qwen3-4B-Instruct-2507.
$\Delta$ is the absolute accuracy change from the base model.
Each final checkpoint uses the same 2,048 examples per benchmark
and greedy decoding.}
\label{tab:forgetting-webshop}
\end{table*}

\section{Model-Compute Accounting}
\label{app:compute_matching}

We compare nominal WebShop training budgets using a
simplified model-compute approximation. Let $P$ denote
model parameters and $L$ a representative processed-token
length per trajectory. All methods initialize the same
base model; the calculation assumes equal $L$ across
methods. 

\paragraph{Counting convention.}
A forward pass contributes approximately $2PL$ and a
backward pass $4PL$. For one optimization epoch, GRPO
requires generation, reference inference, a policy
training forward pass, and a policy backward pass:
\begin{equation}
\begin{aligned}
 C_{\mathrm{GRPO}}
 &= (2+2+2+4)BGPL \\
 &= 10BGPL.
\end{aligned}
\end{equation}
where $B$ is the number of prompts and $G$ the responses
per prompt.

ES evaluates $N$ perturbations on $D$ tasks using
inference only:
\begin{equation}
 C_{\mathrm{ES}} = 2NDPL.
\end{equation}
For $T_{\mathrm{ES}}$ ES updates and
$T_{\mathrm{GRPO}}$ GRPO updates, the nominal full-run
ratio is
\begin{equation}
 \rho =
 \frac{2T_{\mathrm{ES}}ND}
 {10T_{\mathrm{GRPO}}BG}.
 \label{eq:compute-ratio}
\end{equation}

\paragraph{WebShop budgets.}
GRPO uses $B=16$, $G=16$, one optimization epoch per
batch, and 100 updates. Its per-update proxy is
\[
 C_{\mathrm{GRPO}}
 = 10(16)(16)PL
 = 2{,}560PL.
\]
The $16\times16$, $30\times32$, and $30\times100$
ES configurations contribute $512PL$, $1{,}920PL$,
and $6{,}000PL$ per update.

Table~\ref{tab:compute-budgets} summarizes per-update
and full-run ratios.

\begin{table}[H]
\centering
\small
\begin{tabular}{lrrr}
\toprule
ES setup & Updates & Per update & Full run \\
\midrule
$16\times16$  & 200 & $0.20\times$ & $0.40\times$ \\
$30\times32$  & 200 & $0.75\times$ & $1.50\times$ \\
$30\times100$ & 120 & $2.34\times$ & $2.81\times$ \\
\bottomrule
\end{tabular}
\caption{Nominal ES compute relative to GRPO.
The full-run reference is 100 GRPO updates.}
\label{tab:compute-budgets}
\end{table}

\end{document}